\documentclass{article}
\usepackage{ijcai26}

\usepackage{times}
\usepackage{soul}
\usepackage{url}
\usepackage[hidelinks]{hyperref}
\usepackage[utf8]{inputenc}
\usepackage[small]{caption}
\usepackage{graphicx}
\usepackage{amsmath}
\usepackage{amsthm}
\usepackage{amssymb}
\usepackage{booktabs}
\usepackage{algorithm}
\usepackage{algorithmic}
\usepackage{booktabs} 
\usepackage{multirow} 
\usepackage{enumitem}
\usepackage[table]{xcolor}
\usepackage[switch]{lineno}
\usepackage{amsfonts}

\title{\textsc{DeepMed} Search: An Open-Source Agentic Platform for Medical Deep Research with Introspective Verification}

\author{
    Maolin Liu$^1$\and
    Fanyu Xu$^1$\and
    Ruoqing Xu$^1$\and
    Jiahang Zhang$^1$\and
    Hao Wang$^{1,}$\thanks{Corresponding Author.}\And
    Rui Wang$^2$\\
    \affiliations
    $^1$School of Computer Science and Software Engineering, Shanghai University\\
    $^2$Department of Computer Science and Engineering, Shanghai Jiao Tong University\\
    \emails
    \{23721623, 24122195, 50x20, kaito1412, wang-hao\}@shu.edu.cn, 
    wangrui12@sjtu.edu.cn
}

\begin{document}

\maketitle

\begin{abstract}
Navigating the deluge of heterogeneous medical data, from academic literature (PubMed) to clinical guidelines (Web) and private knowledge bases remains a critical bottleneck for evidence-based medicine. While commercial black-box tools lack transparency, standard open-source RAG implementations frequently suffer from ``reasoning drift'' when handling complex, long-tail queries. We present \textbf{\textsc{DeepMed Search}}, a fully open-source, agentic platform designed for transparent medical deep research. Built on a high-performance \textsc{Next.js} architecture, \textsc{DeepMed Search} features a source-adaptive router that autonomously dispatches sub-queries to PubMed, web search, or local graph-based knowledge bases based on information density. Crucially, the platform integrates an introspective verification module, powered by a causal-consistent multi-agent debate framework, to validate retrieved evidence against diagnostic logic before synthesis. To demonstrate its robustness, we showcase \textsc{DeepMed Search}'s ability to autonomously decompose high-difficulty rare disease queries, filter out confounding noise, and generate structured, citation-backed research reports in minutes. By open-sourcing this software, we provide the community with a robust infrastructure to democratize access to trustworthy, glass-box medical reasoning in research and prototyping settings, which is publicly available at: {\url{https://www.deepmedsearch.cloud}} and the demonstration video is available at: {\url{https://youtu.be/4U4aok8yLpk}}.
\end{abstract}
 
\section{Introduction}
 


Large Language Models (LLMs) have revolutionized clinical question answering \cite{singhal2025toward,tu2023generalistbiomedicalai,yang2022gatortronlargeclinicallanguage}, yet their deployment in high-stakes evidence-based medicine remains hindered by limited transparency and auditability \cite{Ji_2023,rudin2019stop}. Medical deep research requires synthesizing heterogeneous data from academic literature, constantly updating clinical guidelines, and structured patient records. While Retrieval-Augmented Generation (RAG) is widely adopted to ground model outputs \cite{lewis2020retrieval,karpukhin2020dense,izacard2021leveraging}, standard RAG implementations often rely on static, linear pipelines that are vulnerable to \textit{Retrieval-Induced Reasoning Drift}, especially in complex, long-tail scenarios such as rare disease diagnosis~\cite{mallen2023not}. In these contexts, standard semantic retrievers suffer from commonality bias \cite{DBLP:conf/icml/KandpalDRWR23}, preferentially surfacing generic evidence that supports common mimic conditions and causing the LLM to be hijacked by plausible but non-discriminative distractors. Although commercial AI research tools attempt to address this issue, their black-box nature prevents clinicians from auditing the causal logic behind retrieved evidence, which remains a non-negotiable requirement in clinical workflows.

To democratize trustworthy and auditable AI in healthcare, we introduce \textsc{DeepMed Search}, a fully open-source, agentic platform engineered for transparent medical deep research. Moving beyond simple RAG frameworks, \textsc{DeepMed Search} introduces a source-adaptive router that autonomously orchestrates sub-queries across heterogeneous endpoints, including PubMed APIs, live web search, and a local Neo4j-based knowledge graph. Furthermore, to ensure the causal alignment and faithfulness of the retrieved evidence, the platform features an introspective verification module \cite{kiciman2024causal}. This engine employs contrastive hypothesis expansion to filter confounding noise and utilizes a multi-agent adversarial debate framework \cite{du2024improving,yao2022react} to rigorously stress-test reasoning paths before final synthesis.

In this paper, we showcase the system's glass-box interactivity, while clinicians can observe \textsc{DeepMed Search} as it autonomously decomposes complex queries, routes searches, and conducts real-time multi-agent deliberations. Ultimately, the system generates structured, citation-backed research reports while maintaining complete interpretability \cite{xiong-etal-2024-benchmarking}.
 
\section{System Architecture}
\begin{figure*}[t]
    \centering
     \includegraphics[width=\linewidth]{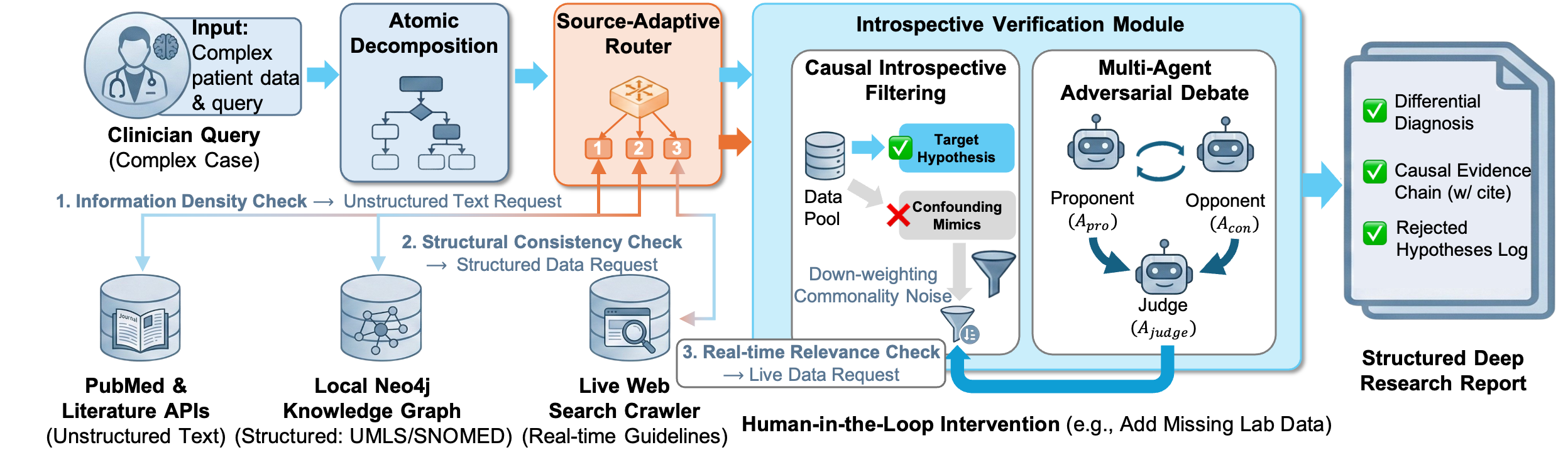} 
    \caption{System architecture of \textsc{DeepMed Search}. The platform integrates a source-adaptive router for multi-endpoint evidence retrieval, causal introspective filtering to mitigate commonality bias, and a multi-agent debate engine to verify reasoning and synthesize reports.}
    \label{fig:architecture}
\end{figure*}

\textsc{DeepMed Search} is engineered as a modular, agentic workflow designed to transform high-level clinical inquiries into structured research insights. As illustrated in Figure~\ref{fig:architecture}, the platform transitions from autonomous data orchestration to rigorous causal verification across three primary stages: Source-Adaptive Routing, Causal Introspective Filtering, and Multi-Agent Synthesis.

\subsection{Source-Adaptive Routing}
A core pillar of \textsc{DeepMed Search} is its ability to navigate heterogeneous data landscapes. Unlike monolithic RAG systems, our platform incorporates a \textbf{Source-Adaptive Router} that treats the information retrieval process as a sequential decision-making task. 

Upon receiving a clinical query, the system first performs \textit{Atomic Decomposition}, breaking the input into distinct clinical constraints (e.g., symptomatic triad, temporal progression, and treatment resistance). The Router then dynamically evaluates the information density of each sub-query to dispatch them to optimal endpoints:
\begin{itemize}
    \item \textbf{Scientific Evidence:} Direct API integration with \textbf{PubMed} for fetching peer-reviewed literature and clinical trial abstracts \cite{qiu2025quantifying}.
    \item \textbf{Pathophysiological Logic:} Querying a local \textsc{Neo4j} knowledge graph (integrating UMLS and SNOMED-CT) to identify multi-hop causal links between symptoms \cite{pearl2009causality}.
    \item \textbf{Real-time Guidelines:} Utilizing a \textbf{Web Search} crawler to retrieve the latest consensus statements and medical news.
\end{itemize}

\subsection{Causal Introspective Filtering}
To mitigate the \textit{Commonality Bias} identified in clinical LLMs \cite{DBLP:conf/icml/KandpalDRWR23}, the platform utilizes a \textit{Causal Introspective Filtering} mechanism. This stage aligns the retrieved evidence with diagnostic logic via two steps: \textbf{1) Contrastive Hypothesis Expansion:} The system proactively formulates ``confounding hypotheses'', prevalent diseases that mimic the target long-tail condition (e.g., generating ``Essential Hypertension'' when analyzing potential ``Pheochromocytoma'') \cite{park2024toward}. \textbf{2) Evidence Down-weighting:} Chunks are filtered based on a discriminative score $S_i = \frac{sim(c_i, H_{target})}{sim(c_i, H_{mimic}) + \epsilon}$. Here, $\text{sim}(\cdot,\cdot)$ denotes the cosine similarity between BGE-M3 sentence embeddings, and $\epsilon$ is set to $10^{-6}$ to avoid division by zero. A candidate chunk is regarded as discriminative when $S_i$ is larger than the threshold $\tau_s$; chunks failing to meet a dynamic threshold are tagged as ``non-discriminative noise'' and visually dimmed in the interface, ensuring the model's attention remains on the long-tail signals.

\subsection{Multi-Agent Synthesis}
The final reasoning phase is handled by an introspective verification module, instantiated as a multi-agent adversarial debate framework \cite{chan2023chateval}.
The system deploys three distinct agents: a \textbf{Proponent} ($A_{pro}$) that constructs reasoning chains from evidence to diagnosis, an \textbf{Opponent} ($A_{con}$) that identifies missing causal links or inconsistencies, and a \textbf{Judge} ($A_{judge}$)  that delivers the final verdict once reasoning converges (Information Gain $< \tau$). 

Crucially, rather than producing a single-line answer, the synthesis engine generates a structured research report. These reports are citation-backed and organized into sections: Differential Diagnosis, Supporting Evidence (from both Graph and Text), and Rejected Hypotheses. This ``glass-box'' output allows clinicians to audit the platform's logic in real-time.

\section{Implementation Details}
The \textsc{DeepMed Search} backend is engineered for high-throughput concurrent processing to support real-time clinical deep research workflows. \textbf{Frontend \& Orchestration:} Built on \textsc{Next.js}, the frontend integrates a \textit{WebSocket}-driven UI for streaming real-time agent deliberations. A dedicated orchestration layer handles the concurrent dispatch of sub-queries via the Source-Adaptive Router. 
\textbf{Knowledge Engines:} Structured multi-hop reasoning is powered by a \textsc{Neo4j} graph database integrating UMLS and SNOMED-CT. For unstructured evidence, we use \textsc{Milvus} to index the MIRAGE corpus (Textbooks and PubMed) \cite{xiong-etal-2024-benchmarking}, utilizing \textsc{BGE-m3} for dense vector similarity coupled with a lightweight cross-encoder for precise re-ranking. 
\textbf{LLM Serving:} To meet the computational demands of deploying three concurrent agents (Proponent, Opponent, Judge), we serve models like \textsc{Qwen2.5-14B} and \textsc{DeepSeek-v3} via \textsc{vLLM}, leveraging \textit{PagedAttention} to optimize memory footprint and maximize throughput.
\textbf{Report Generation Engine:} A custom synthesis module compiles the verified evidence into a structured markdown/PDF report, automatically mapping reasoning steps to specific reference nodes (PubMed IDs or Graph edges) to ensure full traceability.

\section{Demonstration Scenarios}
We present using real-world rare disease cases from the MedR-Bench dataset, as exemplified in Table~\ref{tab:case_study}. The web interface allows clinicians to input complex case reports and interactively audit the deep research process.
\begin{table}[t]
    \centering
    \footnotesize 
    \renewcommand{\arraystretch}{1.1}
    \begin{tabular}{@{}p{0.98\linewidth}@{}}
    \toprule
    \textbf{Query:} \textit{``Patient presents with paroxysmal hypertension, headache, palpitations, and sweating. Standard anti-hypertensive drugs ineffective.''} \\
    \midrule
    \textbf{\textcolor{red}{$\times$ Vanilla RAG Response:}} \\
    Diagnosis: \textbf{Essential Hypertension}. Treatment involves lifestyle changes... \\
    \textit{[Error: Fixates on common disease; ignores paroxysmal \& sweating cues]} \\
    \midrule
    \textbf{\textcolor{blue}{\checkmark \textsc{DeepMed Search} (Ours):}} \\
    Diagnosis: \textbf{Pheochromocytoma}. \\
    \textbf{Reasoning:} 1. \textit{Contrastive Filter:} `Essential Hypertension' down-weighted (cannot explain sweating). 2. \textit{Debate Verdict:} Symptom triad strongly indicates catecholamine-secreting tumor.\\
    \bottomrule
    \end{tabular}
    \caption{Case study: \textsc{DeepMed Search} actively filters common disease noise (Hypertension) to correctly identify the rare condition.}
    \label{tab:case_study}
\end{table}
\subsection{Scenario 1: Autonomous Routing \& Causal Filtering}
Consider a patient presenting with episodic hypertension, headaches, and palpitations. A standard RAG system typically retrieves generic guidelines for ``Primary Hypertension'', leading to reasoning drift. In \textsc{DeepMed Search}: \textbf{1) Routing in Action:} The Router identifies the temporal cue (``episodic'') and dispatches parallel queries to Neo4j and PubMed. \textbf{2) Noise Filtering:} The ``Evidence Panel'' marks generic hypertension evidence with a \textit{Low Relevance} tag (visualized as dimmed text), explicitly stating it was down-weighted due to overlap with common confounders. Conversely, evidence pointing to \textit{Pheochromocytoma} is dynamically highlighted.

\subsection{Scenario 2: Glass-Box Debate \& Report}
\begin{figure}[t]
    \centering
    \includegraphics[width=0.99\linewidth]{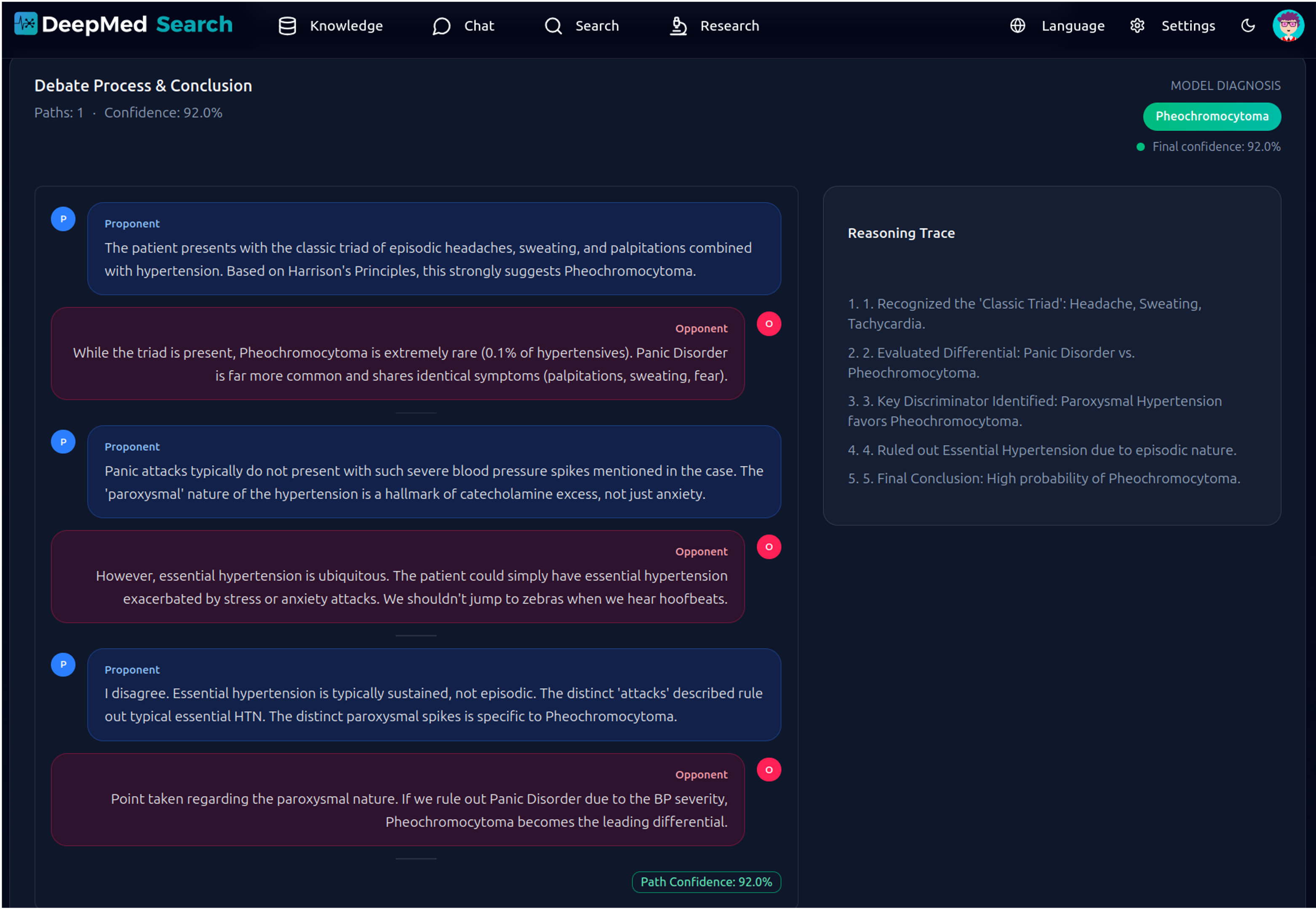}
    \caption{The Multi-Agent Debate Panel. Users can monitor agents as they argue over the diagnosis, providing transparent reasoning logic.}
    \label{fig:interface_debate}
\end{figure}

For complex diagnoses, the clinician can activate the introspective debate mode. \textbf{1) Real-Time Deliberation:} As illustrated in Figure~\ref{fig:interface_debate}, a streaming panel opens where $A_{pro}$ argues for Pheochromocytoma citing paroxysmal symptoms, while $A_{con}$ challenges the evidence density. \textbf{2) Human-in-the-Loop:} Crucially, the clinician can inject missing parameters (e.g., \textit{``Add lab result: elevated metanephrines''}) directly into the debate stream. \textbf{Structured Output:} The $A_{judge}$ finalizes the verdict, and the system instantly compiles a deep research report, linking the diagnosis to verified graph nodes.

\section{Evaluation}
As summarized in Table~\ref{tab:feature_comparison}, \textsc{DeepMed} uniquely integrates source-adaptive routing and multi-agent verification, bridging the gap between vanilla RAG and commercial black-box tools.

\textbf{Performance Gains:} We evaluated on the MIRAGE benchmark \cite{xiong-etal-2024-benchmarking}. DeepMed Search consistently outperforms standard RAG baselines. On the challenging PubMedQA subset, our method improves accuracy by approximately {9.0\%} over vanilla RAG (using Qwen3-8B). For rare diseases, experiments on the MedR-Bench dataset show that our approach achieves an accuracy of {76.43\%} (using DeepSeek-v3.2), bridging the gap between automated systems and expert-level reasoning. 

\begin{table}[h]
    \centering
    \resizebox{\columnwidth}{!}{
    \begin{tabular}{lccc}
    \toprule
    \textbf{Capabilities} & \textbf{Vanilla RAG} & \textbf{Commercial} & \textbf{Ours} \\
    \midrule
    Open-Source Transparency & \checkmark & $\times$ & \checkmark \\
    Adaptive Routing (Web/Graph) & $\times$ & \checkmark & \checkmark \\
    Causal-Consistent Filtering & $\times$ & $\times$ & \checkmark \\
    Multi-Agent Verification & $\times$ & $\times$ & \checkmark \\
    Structured Deep Reports & $\times$ & \checkmark & \checkmark \\
    \bottomrule
    \end{tabular}
    }
    \caption{Feature comparison. \textsc{DeepMed Search} bridges glass-box transparency with practical research-oriented system capabilities.}
    \label{tab:feature_comparison}
\end{table}
\section{Conclusion}

\textsc{DeepMed Search} provides a transparent, agentic platform for medical deep research. It combines source-adaptive routing, causal-consistent filtering, and multi-agent verification to mitigate retrieval-induced reasoning drift in medical LLMs \cite{park2024toward} and generate structured, citation-backed reports. Current limitations include additional computational and latency overhead and the lack of a formal clinician user study. Future work will improve efficiency, involve clinicians in evaluation, and extend the framework to multimodal data such as medical imaging \cite{wang2025medagent}.
\section*{Acknowledgments}
This work was supported by the National Natural Science Foundation of China under Grant No. 62306173.
 
\bibliographystyle{named}
\bibliography{ijcai26}

\end{document}